# Diversity and Inclusion in AI for Recruitment: Lessons from Industry Workshop

Muneera Bano[1], Didar Zowghi[1], Fernando Mourao[2], Sarah Kaur[1], Tao Zhang[2]
[1] CSIRO's Data61, Australia
[2] SEEK, Australia
muneera.bano@csiro.au, didar.zowghi@csiro.au, fmourao@seek.com.au, sarah.kaur@csiro.au, tzhang@seek.com.au

*Abstract*— Artificial Intelligence (AI) systems for online recruitment markets have the potential to significantly enhance the efficiency and effectiveness of job placements and even promote fairness or inclusive hiring practices. Neglecting Diversity and Inclusion (D&I) in these systems, however, can perpetuate biases, leading to unfair hiring practices and decreased workplace diversity, while exposing organisations to legal and reputational risks. Despite the acknowledged importance of D&I in AI, there is a gap in research on effectively implementing D&I guidelines in real-world recruitment systems. Challenges include a lack of awareness and framework for operationalising D&I in a cost-effective, context-sensitive manner. This study aims to investigate the practical application of D&I guidelines in AI-driven online job-seeking systems, specifically exploring how these principles can be operationalised to create more inclusive recruitment processes. We conducted a co-design workshop with a large multinational recruitment company focusing on two AI-driven recruitment use cases. User stories and personas were applied to evaluate the impacts of AI on diverse stakeholders. Follow-up interviews were conducted to assess the workshop's long-term effects on participants' awareness and application of D&I principles. The co-design workshop successfully increased participants' understanding of D&I in AI. However, translating awareness into operational practice posed challenges, particularly in balancing D&I with business goals. The results suggest developing tailored D&I guidelines and ongoing support to ensure the effective adoption of inclusive AI practices.

## I. Introduction

Artificial Intelligence (AI) is increasingly transforming the landscape of online employment marketplaces, with significant potential benefits such as improved efficiency, reduced hiring time, and a broader reach in identifying candidates and job opportunities [1]. AI-powered recruitment systems can streamline resume screening, match candidates to jobs more effectively, and automate parts of the interview process [2]. However, while AI can introduce efficiency, it can also perpetuate harmful biases if not designed and deployed carefully [3]. Models trained on biased datasets may reinforce discrimination, particularly affecting underrepresented groups, resulting in unfair hiring practices and reduced workplace diversity [4]. The balance between leveraging AI's potential and mitigating its risks is a critical area of focus in online employment.

Proactive efforts to promote diversity and inclusion (D&I) are essential to ensure AI-powered hiring tools do not perpetuate or reinforce harmful biases. D&I in AI refers to the proactive incorporation of diverse human attributes and perspectives into all stages of AI development, including design, data, processes, systems, and governance [5]. As per the definition provided by [5], diversity encompasses the representation of individuals with varied attributes protected by law, such as race, gender, disability, and sexual orientation, among others. Inclusion involves ensuring that people with these diverse attributes, particularly those most impacted by AI systems, are meaningfully involved in the design, deployment, and governance of these technologies.

To guide the responsible development of AI systems, including those used in online recruitment markets, researchers at CSIRO's Data61 [1] have developed D&I Guidelines for AI [6]. These guidelines focus on embedding D&I principles into the AI lifecycle, from data collection to deployment, ensuring that AI systems are equitable, transparent, and inclusive. The guidelines advocate human-centred design, diverse stakeholder involvement, and continuous monitoring to address bias and uphold ethical standards in AI applications.

SEEK [2] a market leader in online employment marketplaces across Asia-Pacific (APAC), invests in safe and responsible AI to make employment markets more transparent, efficient and fairer for jobseekers and hirers. As an industry leader, SEEK is committed to promoting diversity and inclusion, recognising that AI systems must be fair and representative to ensure equitable hiring practices. This paper examines the integration of the D&I guidelines developed by researchers at CSIRO's Data61 within SEEK's AI-driven recruitment platform, focusing on how these guidelines can help transform business processes adopted by the industry to effectively deliver fair AI products in online employment markets.

This research investigates the challenge of operationalising D&I guidelines in AI recruitment systems. It aims to identify the challenges, requirements and practices for organisations to integrate D&I in the AI development lifecycle. While awareness of the need for ethical AI is growing, practical frameworks for integrating D&I principles into real-world applications are still lacking [7, 8]. Current approaches rely on theoretical frameworks focused on broad ethical principles and lack practical, industry-led initiatives to ensure the implementation of D&I principles [8]. To tackle this, we conducted a co-design workshop and follow-up reflection interviews in collaboration with SEEK, using AI-generated personas and user stories to explore D&I principles in recruitment scenarios.

The results of this study indicate increased awareness among stakeholders regarding the importance of D&I in AI after the co-design workshop. Key lessons learned include the necessity of tailoring guidelines to specific use cases, the value of stakeholder engagement, and the effectiveness of using generative AI tools to illustrate potential impacts.

---

[1] https://www.csiro.au/en/

[2] https://www.seek.com.au/

This research makes important contributions to the state-of-the-art and practical implementation of D&I in AI-driven recruitment systems. It provides practical and real-world insights into how D&I principles can be operationalised within AI technologies, moving beyond theoretical discussions to actionable, context-specific guidance. The use of co-design workshops with industry stakeholders demonstrated the importance of stakeholder engagement in tailoring D&I guidelines to specific use cases, revealing challenges like balancing business objectives with D&I goals. This study highlights the value of integrating D&I into SEEK's product development and governance workflows, offering a framework to improve inclusivity in their AI systems. Moreover, the results inform the refinement of D&I in AI guidelines and enhance the co-design workshop methodology by emphasising the need for ongoing support, follow-up sessions, and industry-specific adaptations. This work contributes to a broader understanding of how organisations can systematically embed D&I principles in AI systems, and the findings can serve as a model for other industries aiming to ensure inclusive AI-driven processes.

Section II gives background and discuss existing literature on AI in Recruitment, Diversity and Inclusion in AI and Guidelines, and the intersection of these concepts. Section III provides motivation for the research, Section IV Methodology, and Section V describes the results. Section VI provides discussion, Section VII summarises the lessons learned, VIII discusses potential threats to validity, and Section IX draws the conclusion and puts forward the future directions.

## II. BACKGROUND AND EXISTING LITERATURE

### A. AI for Recruitment

AI has the potential to significantly enhance the efficiency and effectiveness of online job seeking and matching processes by automating repetitive tasks, improving discoverability of job opportunities, and enabling data-driven hiring decisions [9, 10]. AI tools can analyse resumes, identify key skills, and match candidates to relevant job opportunities more quickly and accurately than traditional methods [9, 11]. It has been claimed that this results in better hiring outcomes, increased retention rates, and a more streamlined recruitment process [12, 13]. Additionally, AI-based systems can recommend talent lists to hirers, simplifying and accelerating the hiring process [11].

However, significant challenges remain. One of the most pressing concerns is the potential for AI to perpetuate or reinforce harmful biases embedded in historical data, leading to discriminatory hiring practices. A notable example is Amazon's AI-powered hiring tool, which was abandoned after it was discovered to be biased against women [14, 15]. The tool had been trained on resumes submitted to the company over a ten-year period, many of which came from male candidates, leading the AI to penalise resumes from women. This incident illustrates how biased training data can reinforce gender and other discriminatory biases, undermining the fairness of AI-driven recruitment systems [15, 16].

In addition to biases, ethical concerns such as data privacy and transparency present further obstacles to the adoption of AI in recruitment at scale [17]. Moreover, organisational and technological barriers, such as integrating AI into existing hiring frameworks, complicate its widespread implementation [18]. While AI offers significant benefits, ensuring fairness, reducing harmful bias, and maintaining ethical governance are crucial to its effective use in recruitment. Robust guidelines and ongoing monitoring are essential to address these challenges and promote inclusion in AI-driven hiring processes [15]. Employment is considered a high-risk domain for AI in some jurisdictions, leading to proposed regulations that enforce mandatory safeguards to prevent large-scale harm [19].

### B. Diversity & Inclusion in AI

D&I in AI refers to the proactive and intentional integration of diverse attributes and perspectives across all stages of the AI lifecycle, including data collection, processes, system design, and governance [5]. Diversity encompasses attributes protected by law, such as race, gender, age, disability, and more, while inclusion ensures the proactive involvement and representation of diverse individuals impacted by AI systems [5]. Research consistently shows that neglecting D&I in AI development leads to biased algorithms, perpetuating existing societal inequalities and resulting in exclusionary or discriminatory outcomes [7, 20]. For example, AI systems trained on homogenous or biased datasets can disproportionately affect underrepresented groups, with significant implications for sectors like recruitment, healthcare, and law enforcement [21].

The consequences of neglecting D&I principles highlight the importance of understanding the interconnectedness between D&I, equity, and fairness. Equity ensures that everyone, regardless of their background, has equal access to opportunities, while fairness aims to eliminate harmful bias and provide consistent, objective evaluations for everyone. D&I lay the foundation for equitable and fair AI by promoting representation and active participation [7]. When D&I principles are ignored, AI systems can amplify biases, resulting in unfair decision-making processes that erode public confidence and expose organisations to legal and reputational risks [22, 23]. Embedding diverse perspectives throughout AI design helps prevent bias and creates more inclusive, equitable solutions [21].

Integrating D&I into AI systems, especially for online recruitment requires contextualising these efforts to the unique use cases of AI within specific tasks like candidate screening or job matching. We argue that requirements engineering (RE) processes can be tailored to identify and analyse AI-related risks, helping to navigate trade-offs and conflicts that may arise from neglecting D&I principles [24, 25]. For example, RE can support decision-making in situations where maximising inclusion may reduce performance or efficiency, or where ensuring data transparency for under-represented groups might compromise their privacy. By acknowledging diverse users and promoting inclusive system development, RE can help balance conflicting objectives and contribute to the creation of ethical AI systems [24]. Addressing these challenges through an RE lens and operationalising D&I requirements for AI ensures that AI systems are responsibly developed and deployed, promoting a more inclusive, equitable, and ethical AI landscape [25, 26].

*C. D&I Guidelines for AI*

In our research, we utilise the D&I in AI guidelines developed by CSIRO's Data61 [3] which are part of Australia's National AI Assurance Framework [4]. The guidelines are structured around five key pillars: *Humans, Data, Process, System, and Governance.* These pillars provide a holistic approach to embedding D&I principles throughout the AI lifecycle. These guidelines address the critical need to minimise bias, promote fairness, and ensure transparency throughout the AI lifecycle. They provide actionable strategies to tackle challenges such as algorithmic bias, under-representation of diverse groups, and the risk of discrimination in AI decision-making. By incorporating these principles, the guidelines can help AI practitioners navigate the complexities of fairness, privacy, and inclusion, especially in sensitive domains like recruitment, healthcare, and media. To effectively implement these guidelines, co-design workshops with key stakeholders—such as HR professionals, ethicists, legal experts, and underrepresented groups—are vital. These workshops promote collaboration, allowing stakeholders to identify the operationalisable parts of the guidelines that best suit their domain's specific requirements and needs.

*D. D&I in AI for Recruitment*

In AI systems for online recruitment markets, diversity attributes like gender, race, and names are often processed in ways that can either promote fairness or perpetuate harmful bias [27]. Inclusion means that AI fairly evaluates candidates from diverse backgrounds, ensuring equal opportunities for all, while exclusion occurs when AI reinforces existing biases, disadvantaging certain groups in hiring [28].

Several studies highlight that AI in recruitment often exacerbates biases rather than mitigating them, especially when it relies on historical, biased data [28, 29]. Bias in AI algorithms can lead to discriminatory hiring outcomes, particularly against women, racial minorities, and people with disabilities [27, 30]. For instance, studies like [30] illustrate how AI systems, if not designed inclusively, may inadvertently exclude or misjudge applicants with disabilities. Key challenges include the opacity of AI algorithms, which often operate as "black boxes," and the lack of transparency around how AI systems make decisions [27, 29].

In addressing these issues, AI ethics guidelines such as those proposed by IEEE and the European Union call for fairness, transparency, and accountability in AI systems. However, many of these guidelines, as seen in studies like [29], primarily focus on broad ethical principles without a dedicated emphasis on D&I. Furthermore, there is often little effort to engage in *industrial co-design activities*, where stakeholders, including those from underrepresented groups, collaborate to ensure D&I principles are actively integrated into AI development [31, 32]. Most studies, including those on AI recruitment strategies, rely on theoretical frameworks and literature reviews but lack practical, industry-led initiatives to ensure continuous implementation of D&I principles [33].

There is clear evidence that AI systems are not free from bias [5], and their unchecked use in job matching and recruitment can lead to discriminatory outcomes [29]. While the existing literature has made efforts to incorporate ethical AI standards, none have focused specifically on D&I in AI guidelines. This leaves a critical gap in ensuring fairness in AI-driven recruitment systems.

## III. RESEARCH MOTIVATION

The lack of real-world validation and co-design activities remains a critical gap. Without involving industry stakeholders and ongoing collaboration, AI tools in recruitment will likely continue to reinforce existing biases, despite adherence to ethical AI guidelines. Addressing these gaps through practical engagement and industry validation is essential for building AI systems that are truly inclusive and capable of enhancing diversity in the workplace [34, 35].

There is a pressing need to explore how D&I guidelines can be applied in the context of AI for online recruitment markets. Doing so would help mitigate biases and promote inclusivity, ensuring that AI systems operate fairly for all candidates and hirers. To the best of our knowledge, none of the studies in the literature have involved industry partners in a co-design workshop to create contextualised guidelines for operationalising D&I principles.

This motivated us to explore and conduct a co-design workshop with SEEK, a leading online employment marketplace in APAC, serving millions of job seekers and hirers. SEEK is at the forefront of digital recruitment, making it an ideal partner to examine how D&I guidelines in AI can be operationalised in real-world scenarios. By collaborating with a large organisation like SEEK, which integrates AI into its recruitment processes, we aimed to ensure that AI-driven hiring is inclusive, and free from harmful bias.

This research investigates two research questions to identify the challenges, requirements and practices required for SEEK to integrate the D&I guidelines in the AI development lifecycle:

> **RQ 1:** *How can D&I in AI guidelines be tailored to specific industry sectors to provide targeted and practical guidance for practitioners?*

Customising D&I guidelines for specific sectors like recruitment is crucial because AI tools need to reflect the unique needs and dynamics of each industry. In recruitment, bias in AI can lead to discriminatory hiring practices, impacting underrepresented groups. By tailoring guidelines, we can ensure they address these sector-specific challenges and offer actionable insights for AI-driven recruitment.

> **RQ 2:** *How can AI professionals be supported in applying D&I in AI guidelines more effectively within their unique contexts?*

AI professionals often face challenges in applying ethical guidelines due to the complexity and variability of real-world scenarios. Providing sector-specific tools, processes, training, and support will enable them to embed D&I principles more effectively, ensuring AI systems operate fairly and inclusively.

---
[3] https://research.csiro.au/ss/guidelines-for-diversity-and-inclusion-in-artificial-intelligence/

[4] https://www.finance.gov.au/government/public-data/data-and-digital-ministers-meeting/national-framework-assurance-artificial-intelligence-government

## IV. METHODOLOGY

At the outset of the workshop, the research team sought approval from human research ethics for industry collaboration, data collection, analysis, and publication of the results. The research design was divided into three phases: pre-workshop preparation, activities during the co-design workshop, and post-workshop reflection interviews. Besides detailing each phase, this section outlines the data collection and analysis processes used to assess the workshop's outcomes.

### A. Pre-Workshop Preparation

In the pre-workshop phase, extensive preparation was undertaken to ensure the workshop was focused and productive on identifying opportunities for systematic D&I in AI improvements in prioritised use cases. A preliminary meeting was held with SEEK's Head of Responsible AI (RAI) to identify key AI use cases where D&I guidelines could be integrated.

Two AI-driven recruitment use cases were identified for incorporating D&I guidelines. These use cases represent areas where AI plays a critical role in decision-making, and thus, integrating D&I principles would uplift existing mechanisms to ensure fairness and reduce bias. Below are descriptions of the use cases:

*1) Strong Applicant Badge*

The "Strong Applicant Badge"[5] use case is an AI-driven feature designed to inform jobseekers when they might be a strong fit for an advertised role. Predictions are based on matching criteria between each job advertisement description and factors extracted from the jobseeker's resume or profile at SEEK, such as skills and work experience. The badge is meant to help jobseekers gain confidence to apply for certain jobs. This use case presents potential D&I challenges, as the criteria used to assess compatibility might inadvertently favour certain groups while disadvantaging others if biased data or business decisions are used. Incorporating D&I guidelines into this feature would strengthen existing RAI guardrails that ensure the AI model fairly represents diverse candidates and avoids reinforcing harmful biases.

*2) Ads Summarisation*

The "Ads Summarisation" use case involves the automatic generation of summaries for job advertisements. This AI tool aims to reduce the time and effort required by recruiters to create concise, effective job ads that attract the right candidates. D&I considerations are critical in this process, as the language and tone of job ads can subtly influence which demographics apply for a role. By applying D&I guidelines, we can better determine to what extent the AI Ads summarisation tool was designed to create summaries that are inclusive and free from biased language, ensuring that all potential candidates feel welcome to apply.

Once the use cases were defined, the research team utilised a tailored user story template. The research team generated user stories based on the confirmed use case descriptions. These user stories were then structured according to the five pillars of the D&I in AI guidelines. Additionally, the research team reviewed the D&I in AI guidelines and mapped the user stories against the specific guidelines (see Appendix B). The user stories reflected diverse roles representing various attributes, such as gender, ethnicity, and age, to help workshop participants imagine the potential D&I impacts of AI in job-seeking processes.

### B. Co-Design Workshop

The co-design workshop was held at SEEK's Melbourne Head Office. The participants included product owners, data scientists, and RAI governance experts who have worked in at least one of the selected use cases. The following key activities were conducted during the workshop:

1. **Introduction and Use Case Presentation**: Each session began with a detailed presentation of one of the selected use cases. These included descriptions of the AI models, workflows, and the ethical challenges related to D&I. The presentations aimed to provide participants with a clear understanding of how D&I issues might arise in these AI-driven recruitment processes.

2. **User Story Generation**: Participants were guided through the process of generating their own user stories, using a structured template. These user stories were designed to identify potential D&I challenges and offer solutions. Pre-generated user stories by the research team were used as a starting point, helping participants consider how the D&I guidelines could be applied within each use case. In software development and product management, a user story is an informal, natural language description of one or more features of a software system. User stories are part of the Agile approach that helps shift the focus from writing about requirements to talking about them [36]. These stories typically follow a simple template:

   > *As a [type of user], I want [an action] so that [a benefit/value].*

   We adopted the tailored user story template for eliciting and capturing D&I in AI requirements proposed by [25]. The template particularly focuses on roles embodied by persons with diverse attributes or that specify diverse system behaviours. The template we utilised for D&I-related user stories is the following:

   > *As a <diverse role>, I want <action | process | artifact> to <predicate | behaviour> so that <rationale>*

   This format helps to identify the key stakeholders, the desired action or outcome, and the underlying motivation behind the request. By employing this user story template, the development team can better address the needs of diverse stakeholders and align their AI systems with the overarching goals of diversity and inclusion.

3. **Diverse Roles**: Throughout the workshop, participants worked with diverse roles representing various demographic attributes, including gender, ethnicity, and age. This approach enabled participants to explore how

---

[5] https://help.seek.com.au/s/article/What-is-the-You-might-be-a-strong-applicant-badge

AI systems could impact individuals from different backgrounds, ensuring that the D&I guidelines were applied to a wide range of user scenarios.
4. **Prioritisation of User Stories**: After generating the user stories (See Figure 1), participants engaged in a group voting process to prioritise which stories should be focused on first. This prioritisation helped determine which identified D&I features or concerns were most important to address in SEEK's AI systems, guiding future development efforts.

*C. Post-Workshop Evaluation*

Following the workshop, the user stories generated by participants were further analysed by the research team. These user stories were mapped back to the D&I in AI guidelines to ensure that the guidelines were effectively tailored to SEEK's specific recruitment context. This phase also included post-workshop discussions on how the guidelines could be practically integrated into SEEK's existing AI systems and governance processes.

Finally, a post-workshop reflection was conducted with participants through follow-up interviews approximately after six months. These interviews evaluated the long-term impact of the co-design workshop, gathering feedback on the effectiveness and practical utility of the D&I guidelines. This reflection phase provided critical insights into how well the guidelines were received and considered for embedding into SEEK's AI system's development process and helped identify areas for improvement.

*D. Data Collection and Analysis*

Data collection for this study occurred during two key phases: during the co-design workshop and in the post-workshop evaluation. During the workshop, data were collected in the form of:

- **User Stories**: Generated by the participants to explore D&I challenges and solutions.
- **Diverse Roles**: Insights from participants regarding how the diversity of roles in user stories influenced their understanding of D&I in AI systems.
- **Voting and Prioritisation Results**: Participants' votes on the user stories provided data on which D&I concerns were most important to address.

In the post-workshop phase, additional data was collected through interviews. Approximately six months after the workshop, semi-structured interviews were conducted with participants to gather feedback on the long-term effectiveness of the D&I guidelines and their consideration into SEEK's AI systems. The interviews were structured around four key themes: *awareness*, *operationalisation ability*, *team dynamics*, and *real-world impact* (see Appendix A for interview questions). For each theme, participants were first asked a closed-ended question to quantify their response, followed by an open-ended follow-up question to gather detailed insights. The interviews were conducted via video calls and recorded with participants' consent. A thematic analysis was performed on the qualitative responses, identifying common themes related to D&I awareness, operationalisation, and team dynamics. Quantitative data from the closed-ended questions were summarised to provide a general measure of the workshop's impact. Direct quotes from participants were extracted to illustrate key findings and provide personal perspectives on the workshop's outcomes.

The additional benefit of this analysis was that it was helpful in refining and improving the D&I in AI guidelines and ensuring they were aligned with the specific needs of SEEK's recruitment processes. The interviews conducted approximately six months after the workshop were analysed to assess the ongoing integration of D&I guidelines into SEEK's systems. This analysis provided insights into the effectiveness of the co-design approach and highlighted areas where further improvements were needed.

V. RESULTS

*A. Co-Design Workshop*

The workshop allowed participants to discuss the use cases, generate user stories, and elaborate the discussion around D&I concerns and improvements within SEEK's AI tools. Figure 1 shows a snapshot of Miroboard screen capturing the diverse user stories generated by participants of a co-design workshop. These stories reflect various roles and highlight specific requirements for improving diversity and inclusion in AI-driven recruitment processes. By synthesising these stories, key requirements emerge for making recruitment more accessible and inclusive across a wide range of candidates. Key roles that emerged include hiring managers, job seekers with specific needs (such as non-native English speakers, people with disabilities, and those requiring visa sponsorship), as well as AI leaders responsible for shaping more inclusive recruitment systems.

In terms of the aspects of D&I, accessibility was a major concern, with stories from people with disabilities and non-native English speakers calling for job ads that clearly provide relevant information, ensuring these groups can engage meaningfully with job opportunities. Personalisation was another significant theme, as job seekers expressed the need for AI to tailor job summaries to their individual needs, streamlining their search process. Furthermore, bias mitigation was a recurring topic, with AI leaders emphasising the need for systems that reduce human biases in job ad creation, ensuring fairness across diverse candidates, including gender, race, language proficiency, and disability.

Below are the key findings from the workshop, structured around the two selected use cases along with SEEK's governance processes.

*1) Strong Applicant Badge Use Case*

The Strong Applicant Badge, which flags job seekers as highly suitable for specific roles based on AI-driven analysis, was discussed in depth. Participants reinforced several key D&I concerns:

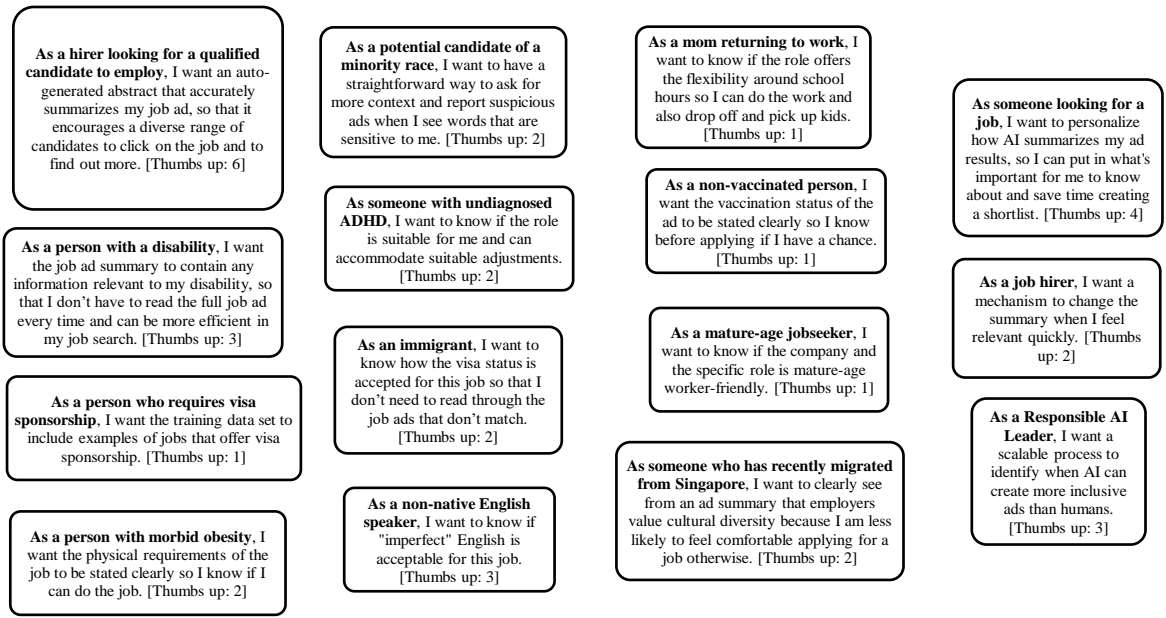

Figure 1: Some of the User Stories from Co-design workshop Miro board and interactive prioritisation for the Ads Summarisation Use Case.

- **Representative Data Samples**: Participants highlighted the challenges of gathering representative training data for minority groups, as collecting protected attributes in the hiring process is legally sensitive. Although organisations can request jobseekers this data for specific research, small sample sizes and the bias of self-reporting remain significant obstacles.
- **Fairness Assessment**: The lack of access to protected attribute data limits organisations' ability to accurately measure diversity and promote inclusion. Participants also discussed the ethical use of inferred data and its limitations in fairness assessments.
- **Explainability and Transparency**: The importance of both internal and external transparency around AI models and their driving factors was emphasised as another key consideration, especially given that latent variables may unknowingly influence the model's behaviour and outcomes.
- **Cognitive Bias**: The impact of seeing (or not seeing) a Strong Applicant Badge was briefly discussed, highlighting the psychological and cognitive bias influence this could have on a candidate's decision to apply for a job.

**Recommendations**: After reflecting on the key D&I concerns and comparing them with existing guidelines, participants were asked to provide actionable recommendations for project leaders to enhance D&I efforts with specific goals in mind. Key recommendations included:

- Expanding the monitoring of badge visibility beyond gender to assess its impact on minority applicants, including First Nations candidates.
- Conducting additional stakeholder engagement surveys to better understand how minority groups perceive job applications and their confidence in applying.

*2) Ads Summarisation Use Case*

The second use case focused on the automation of job ad summaries. In this case, participants highlighted the following D&I concerns:

- **Efficient Human Oversight**: Accuracy and false positive rates are crucial for scalability and efficient human resource allocation. Job ads flagged as sensitive by bias detection mechanisms can lead to high and inefficient manual efforts in curating summaries if not carefully managed.
- **Cultural and Linguistic Bias**: The discussion highlighted that language models often struggle with gendered languages (e.g., French), making it more challenging to create neutral summaries for roles that may be linguistically gendered in some languages.
- **Biased Ground Truth**: The AI system is evaluated based on how well it replicates human-generated ad summaries, which means it can unintentionally reinforce unconscious human biases.

**Recommendations**: Participants reflected on the identified D&I concerns and proposed the following key recommendations for project leaders:

- Enhancing prompt engineering to incorporate stronger D&I considerations.
- Leveraging synthetic training data that includes both "good" and "bad" examples to help mitigate human bias.

The workshop effectively raised awareness and fostered meaningful discussions around D&I in AI. However, challenges like limited resources and unclear guidelines still hinder the full integration of D&I principles in SEEK's systems.

*B. Post Workshop Reflection Interviews*

This section synthesises feedback from eight SEEK participants who attended the workshop (and consented to being interviewed) aimed at enhancing their awareness and ability to implement D&I in AI within their product design and governance workflows. The workshop took place approximately six months prior to these interviews, and the purpose of these interviews is to evaluate its long-term impact based on four key themes as follows:

*1) Awareness of D&I in AI Principles*

Six out of the eight participants reported an increase in their awareness of D&I in AI principles because of the workshop. Participants generally attributed this increase to the dedicated time spent reflecting on these issues and engaging in facilitated discussions. Several noted that the workshop allowed them to think more deeply about the impacts of AI on different stakeholders.

For example, one participant noted that the workshop provided valuable time to reflect on D&I, which they rarely do in their regular work. Another participant mentioned how their team expanded the concept of fairness to incorporate D&I principles in their service evaluations after the workshop. Additionally, one participant appreciated the diverse perspectives offered by external facilitators (i.e. researchers), which deepened their understanding of D&I considerations.

However, some participants had difficulty recalling specific details of the workshop. One participant mentioned that they had an "aha" moment during the workshop but could not recall the exact insight that increased their awareness. Another participant pointed out that, although they already had a high level of awareness of Responsible AI prior to the workshop, the session further reinforced their understanding of D&I by highlighting new perspectives.

While individual awareness increased for most participants, a few noted that prioritising D&I in their daily activities is challenging. One participant mentioned that while their awareness had increased, the practical application of D&I principles within their team remained limited, as satisfying other functional requirements took precedence.

*2) Ability to Operationalise D&I in AI*

**Limited Improvement in Operationalisation:** While six participants reported increased awareness, only three felt that their ability to operationalise D&I principles had improved. Several participants expressed challenges in translating the workshop's concepts into actionable steps within their workflows. For instance, one participant stated that while they were aware of the discussions around mitigating D&I risks, they were unclear on how to enforce these strategies in practice. Another participant highlighted that their team struggled to balance the operationalisation of D&I principles with business priorities, especially when these principles did not directly translate to revenue generation.

**Success in Specific Areas:** There were a few instances where operationalisation was successful, for example one participant mentioned that their team had begun to frame user questions around D&I in ways they wouldn't have considered prior to the workshop, though broader operationalisation challenges remained.

**Barriers to Operationalisation:** Several barriers to operationalising D&I principles were identified. A common issue was the lack of demographic data needed to measure D&I impacts on minorities. Time and business pressures usually emerge as significant barriers in practice. One participant mentioned that while their team had identified potential D&I concerns, determining prioritisation to better understand these concerns was a challenge.

Additionally, none of the participants could recall seeing the workshop output notes and recommendations after the workshop, this may have contributed to the challenge of lack of sustained support or follow up to take action, after gaining awareness and insights for increasing D&I considerations.

*3) Team Dynamics*

**Endorsement of D&I Rituals:** Most participants (seven out of eight) indicated that they would endorse the introduction of a clear, guided business process for D&I guidelines at SEEK. One participant explained that their team already applied fairness principles that aligned with D&I, and they would support formalising these efforts into a broader process. Another participant emphasised the importance of upskilling SEEK's RAI champion within each team on D&I guidelines to initiate regular reviews of projects, ensuring that D&I principles are consistently considered. This aligns with the D&I in AI guidelines, which emphasise the crucial role of AI stewards in maintaining accountability and promoting ethical AI practices across teams[6].

**Team Engagement and Challenges:** Despite this endorsement, actual changes in team dynamics post-workshop were limited. Many participants indicated that they had not observed significant shifts in their teams' approach to operationalising D&I. One participant mentioned that, although they valued the workshop, they hadn't seen practical changes within their team. Another participant noted that while they would support the introduction of formal D&I processes, the perception that such processes are time-consuming could be a barrier to widespread adoption. A speculative reason for these challenges could be that implementing practical changes to existing business processes, or altering team structures, requires strong enforcement and clear direction from leadership. Without explicit management endorsement, such as incorporating D&I practices into standard business processes or mandating compliance as part of Responsible AI governance, teams may not feel the urgency or authority to prioritise these changes

**Team Engagement with D&I Rituals:** Some participants observed a lack of engagement from their teams. One participant noted that while they and another colleague were keen to adopt D&I principles, some of their team members viewed RAI training as a burden and were less enthusiastic about engaging with D&I concepts. This highlights the need for continued RAI and D&I culture strengthening among technical teams. This feedback also entails the necessity of broader team involvement and possibly more engaging formats to encourage participation in D&I rituals.

---

[6] https://research.csiro.au/diai/role_taxonomy/ai-ethicist-responsible-ai-lead-steward/

*4) Challenges for SEEK*

Despite having a mature RAI framework, RAI training and processes to assess and monitor AI systems along AI principles, including fairness, the workshop results highlighted D&I operationalisation challenges for:

**Continuous Engagement:** While participants (five out of eight) felt that the workshop contributed positively to D&I discussions within SEEK, specific, tangible, and long-term improvements require continuous engagement. One participant mentioned that the workshop helped to provide a systematic view of how D&I could be incorporated into the AI lifecycle, allowing their team to think more comprehensively about these issues. Several participants suggested that ongoing workshops or follow-up sessions could help maintain the momentum from the initial workshop. One participant proposed having more explicit follow-up sessions to ensure that discussions lead to concrete actions. Another participant mentioned the importance of involving more people across teams, especially those less familiar with D&I, to broaden the workshop's impact.

**Measurable Progress:** Despite the positive reflections on the workshop's impact, many participants did not see significant tangible outcomes. One participant noted that they could not recall specific actions taken as a result of the workshop. Another participant mentioned that while D&I principles were being discussed within SEEK, they could not remember any direct outcomes or improvements stemming from the workshop itself.

*C. Insights from SEEK's RAI Leadership*

In addition to conducting post-workshop reflection interviews, we also sought insights from SEEK's Responsible AI leadership to gather their thoughts on the effectiveness of the workshop and how D&I principles are being integrated into ongoing AI practices within the organisation.

SEEK's RAI Leadership highlights that the sustainable integration of D&I guidelines into AI governance is a critical milestone for organisations aiming to enhance D&I maturity. This requires a systematic approach, ensuring that D&I principles are embedded in business processes and RAI frameworks rather than being treated as isolated initiatives. A key insight is that D&I workshops alone are insufficient for cultural transformation; the change must permeate organisational culture to be sustainable.

Moreover, RAI leadership emphasises the challenge of balancing multiple AI ethics guidelines, noting that it is impractical to apply all guidelines in every use case. Organisations should focus on quickly identifying high impact use cases that require deeper D&I considerations. By doing so, they can prioritise D&I efforts where they will have the most significant effect, ensuring that AI systems are not only fair but also human-centric. Additionally, organisations should go beyond risk management and unlock the potential benefits that D&I for AI can bring.

To operationalise these insights, actionable recommendations include consolidating a database of context-specific user stories to guide D&I discussions, updating AI Impact Assessment processes to integrate D&I, and institutionalising co-design workshops. Engaging external D&I experts can also bring fresh perspectives and ensure continuous improvement. These steps will help organisations foster a more inclusive, ethical AI landscape while maximising the potential benefits of D&I in AI.

## VI. DISCUSSION

The results from the co-design workshop and post-workshop reflection interviews provide valuable insights into the operationalisation of D&I in AI systems within the online employment context, particularly in SEEK's AI-powered platform. The key takeaway is an increased awareness among participants about the challenges of embedding D&I principles into AI workflows, but the results also highlight significant obstacles in translating this awareness into actionable change. These findings offer both confirmations of existing literature and new insights into how D&I principles can be more effectively implemented in real-world scenarios.

*A. Increasing Awareness of D&I Challenges in AI*

The workshop and subsequent interviews demonstrate that participants gained a clearer understanding of the complexity of D&I issues in AI-driven recruitment, particularly in terms of bias detection, data limitations, and the ethical considerations around transparency and explainability. This increased awareness reflects findings in the broader literature that highlight AI's potential to exacerbate biases if not carefully managed [27]. For example, participants recognised the challenge of lacking access to protected demographic attributes such as race or ethnicity, which makes evaluating the D&I specific performance of models difficult. This aligns with prior research that identifies data availability as a critical factor in addressing AI bias [30].

*B. The Importance of Contextualised Guidelines*

One of the new insights offered by this research is the importance of tailored D&I interventions for specific use cases, such as the Strong Applicant Badge and Ads Summarisation. The workshop revealed that general D&I guidelines, while helpful in raising awareness, are insufficient for addressing the nuanced challenges that arise in specific AI-driven tasks. Participants recognised the need to contextualise guidelines to ensure they are applicable to the unique objectives of each AI tool. This finding extends the current literature, which often focuses on broad, ethical AI standards but does not emphasise the need for customised, context-sensitive guidelines for different AI use cases [28].

For example, in the Strong Applicant Badge use case, the challenge of balancing merit-based selection with diversity goals surfaced, showing that the generic fairness principles in AI ethics literature cannot easily address the intricacies of promoting diversity without sacrificing perceived merit [37]. This confirms that contextualised guidelines tailored to the task at hand are essential to operationalising D&I in a meaningful way. Similarly, in the Ad Summarisation use case, the challenge of adjusting language to comply with international D&I standards required not only ethical considerations but also legal and cultural sensitivity. This complexity shows the importance of developing guidelines that are not only ethical but also feasible and relevant to the specific tasks AI systems are designed to perform. Contextualised guidelines allow teams to focus on the areas where D&I risks are highest, such as language models or candidate ranking systems, and integrate D&I into the workflow in a practical, effective manner.

*C. Comparison to Existing Literature*

The outcomes of the workshop also provide a valuable comparison to existing literature on AI ethics and recruitment. Previous research has established that AI systems in hiring often perpetuate biases due to their reliance on historical data [29]. The workshop outcomes highlighted this risk, with participants acknowledging that AI models, especially when delivered as "black boxes," can perpetuate historical biases present in the training data. Participants also emphasised the need to move beyond the typical focus on bias detection and algorithmic fairness, advocating for the use of D&I guidelines to actively promote greater inclusion.

The workshop's use of diverse roles in user stories to simulate the impact of AI on different user groups also aligns with literature advocating for human-centred design in AI systems [38]. By exploring user stories from the perspective of various demographic groups, participants were able to identify potential bias risks that might not have been visible through a purely technical lens. This approach echoes calls in the literature for integrating D&I at the design stage of AI systems [25, 26].

*D. Application to Other Use Cases in Recruitment*

The insights drawn from the workshop have broader applications beyond the specific use cases discussed. The challenges identified in embedding D&I principles into AI models for recruitment are likely to apply to other areas of online employment, such as resume screening or automated interview systems. These AI systems, which increasingly rely on predictive algorithms, face similar risks of perpetuating bias if D&I considerations are not thoroughly embedded during design and development.

*E. Addressing the Research Questions*

Our research design integrates co-design workshops, user story templates, and post-workshop reflection interviews, offering a structured approach to developing tailored guidance for D&I in AI. This method ensures practical, real-world insights and actionable steps to effectively embed D&I principles in AI-driven systems.

---

**RQ 1:** *How can D&I in AI guidelines be tailored to specific industry sectors to provide targeted and practical guidance for practitioners?*

D&I guidelines for AI systems must be customised to specific industry sectors by aligning them with the particular challenges, regulations, and objectives of each domain. For example, in the recruitment industry, guidelines need to address issues such as bias in candidate selection algorithms, fair representation in job ad summaries, and ensuring inclusivity in automated hiring processes. Customisation can be achieved by working closely with industry stakeholders to identify relevant use cases, developing essential user stories, and designing guidelines that incorporate the specific ethical, legal, and operational requirements of the sector. This ensures that D&I principles are not generic but directly applicable to the unique tasks AI systems are designed to perform.

---

To effectively tailor and apply D&I guidelines in AI, it is essential to customise them for specific industry needs while providing AI professionals with the practical tools and support required to integrate these principles into their unique contexts.

---

**RQ 2:** *How can AI professionals be supported in applying D&I in AI guidelines more effectively within their unique contexts?*

To help AI professionals effectively apply D&I guidelines, it is essential to provide them with practical tools, frameworks, and ongoing training. This includes developing clear, actionable steps for embedding D&I principles into AI design, implementation, and governance processes. Co-design workshops, such as the one conducted in this study, offers hands-on experience and collaborative learning opportunities, allowing professionals to explore real-world use cases and understand the potential D&I impacts of their AI systems. Additionally, providing continuous support through follow-up sessions, resources, and internal champions can ensure that professionals are equipped to apply D&I guidelines consistently within their specific contexts. Also, the directives and encouragement from higher levels of management within the organisation would play a significant role in enforcing these practices in the revised workflows.

---

VII. LESSONS LEARNED

By synthesising the data from both the workshop and the interviews, we identified recurring themes and areas of improvement, leading to the formulation of lessons that can inform future efforts in integrating D&I into AI development and governance. These lessons offer practical, industry-specific insights into the challenges and opportunities of implementing D&I principles in AI systems, providing valuable guidance for other organisations looking to ensure their AI-driven processes are fair, inclusive, and aligned with real-world business needs. We summarise the lessons learnt below.

- **Contextualising D&I guidelines for specific AI use cases is key to successful operationalisation**, as generic guidelines are challenging to apply effectively without tailored approaches for individual tasks or tools.
- **Operationalisation efforts benefit from the use of diverse roles and user stories**, as observed in the co-design workshop, which allowed participants to better understand the real-world impacts of AI on different demographic groups.
- **Industry-specific D&I strategies should be co-designed with stakeholders** to ensure that practical, actionable insights are generated and incorporated into AI systems, allowing for the effective alignment of D&I principles with operational goals.
- **Sustained engagement post-workshop is necessary to reinforce D&I learning and apply new insights effectively**, ensuring D&I principles are continuously integrated into daily workflows.
- **Translating D&I awareness into practical action is challenging due to business priorities, lack of demographic data, and unclear guidelines**, emphasising the need for contextualised guidance that aligns with specific use cases.
- **The lack of access to protected demographic data makes it difficult to evaluate AI models' D&I performance and detect biases**, necessitating privacy-compliant data collection methods for evaluating and improving D&I outcomes.
- **Team engagement and leadership are crucial for operationalising D&I, and formal processes or dedicated roles may be required**, such as appointing RAI champions within teams and upskilling them on D&I to advocate and monitor progress.

- **Balancing D&I concerns with business goals requires clearer, context-specific guidance**, making co-design workshops important for developing practical solutions that do not compromise either D&I objectives or business performance.
- **Clear accountability mechanisms and regular reviews are essential for embedding D&I principles into AI governance frameworks**, ensuring consistent oversight and evaluation of D&I efforts in AI systems.

## VIII. Threats to Validity

For this study, which utilised an exploratory co-design workshop and post workshop reflection interviews, we identify and address threats to internal, external, and construct validity as follows:

### A. Internal Validity

Internal validity refers to the degree to which the results of the study accurately reflect the intended phenomena, without being influenced by extraneous variables. A primary internal validity threat in this research is the participant bias during the co-design workshop. As the workshop participants were SEEK employees who were involved in the development or management of the AI systems, their familiarity with the products might have influenced their responses. Additionally, social desirability bias may have affected participants' feedback in the interviews, as they may have felt inclined to give more positive feedback about the workshop and their subsequent awareness of D&I issues. To mitigate this, the exploratory nature of the research involved open-ended, reflective discussions during the workshop and follow-up interviews, encouraging participants to provide honest feedback, including critiques and challenges faced. The involvement of external facilitators from outside SEEK (research team) also helped create an environment where participants could discuss both positive and negative aspects of the AI systems.

### B. External Validity

External validity concerns the extent to which the results of this study can be generalised to other contexts or settings. Since this study was conducted with SEEK, a specific company operating in the online employment sector, the findings may not be directly applicable to other industries or companies with different AI systems, RAI governance or organisational structures. The specificity of the use cases, such as the Strong Applicant Badge and Ad Summarisation, further limits the generalisability to AI systems used in other areas of online recruitment.

However, by focusing on industry-specific use cases and exploring D&I principles in real-world AI applications, this study offers insights that can guide future research and implementation in other industries. While the results may not be fully generalisable, the lessons learned around contextualising D&I guidelines and addressing bias are relevant to other organisations seeking to implement ethical AI systems.

### C. Construct Validity

Construct validity addresses the accuracy with which the concepts or phenomena being studied are represented by the research design and methods. One threat to construct validity is whether the D&I concepts were understood and applied consistently by all participants during the workshop and interviews. Since D&I can be interpreted differently across teams and individuals, there is a risk that participants may have focused on narrower aspects of D&I (e.g., gender diversity) rather than a holistic view that includes various dimensions like ethnicity, disability, and socioeconomic status.

To address this, the research design included pre-workshop materials and facilitated discussions that helped define and standardise the key D&I concepts. The use of personas and user stories during the workshop also provided a structured way for participants to engage with D&I principles in relation to their specific AI use cases. Furthermore, the follow-up reflection interviews provided an opportunity to clarify participants' evolving understanding of D&I and to assess the broader impacts on their team dynamics and operational practices.

## IX. Conclusion and Future Work

This research explored the application of D&I principles in AI-driven recruitment systems through a co-design workshop conducted with SEEK, followed by post-workshop reflection interviews. The findings indicate that while the workshop successfully increased participants' awareness of D&I issues, translating this awareness into practical and operational changes remains challenging. Key barriers include a lack of access to demographic data, difficulties in balancing D&I principles with business priorities, and the need for clearer guidelines tailored to specific use cases. The study highlights the importance of contextualised D&I guidelines for AI systems, as generic ethical AI principles are not easily embedded into the specific operational tasks of recruitment platforms. The research further highlights the need for ongoing support, follow-up activities, and structured processes to ensure that D&I principles are integrated into AI development and governance effectively. (also mention the role of leadership and management)

In future work, we aim to develop more specific, actionable D&I guidelines that are tailored to the unique requirements of various AI use cases, extending beyond the recruitment sector. We plan to conduct longitudinal studies to assess the long-term impact of D&I principles on hiring outcomes and diversity metrics, as well as explore privacy-compliant methods for collecting demographic data to evaluate AI systems' performance in this area. Additionally, we will focus on integrating D&I considerations into AI governance frameworks, ensuring these principles are consistently applied throughout the AI lifecycle. These efforts will help us operationalise D&I more effectively in AI-driven systems, leading to fairer and more inclusive recruitment processes and beyond.

From a research perspective, conducting this research with industry partners has helped us refine and improve our method for tailoring our D&I in AI guidelines. The revised method will be used with future industry collaborators from different sectors.

## Acknowledgment

We would like to express our appreciation to the management and staff of SEEK for their participation in this collaborative research project.

## APPENDIX A – POST WORKSHOP REFLECTION INTERVIEW QUESTIONS

The interviews were structured around four key themes: **awareness**, **operationalisation ability**, **team dynamics**, and **real-world impact**. For each theme, participants were first asked a closed-ended question to quantify their response, followed by an open-ended follow-up question to gather detailed insights. This mixed-method approach allowed for both measurable data and nuanced reflections.

**Awareness**
1. Participants were asked whether their awareness of D&I in AI principles had increased after the workshop with the prompt:
    - "My awareness of D&I in AI principles increased after the workshop." (True/False)
2. **Follow-up**: "In what ways has your awareness increased?"
    - This question explored how the workshop impacted their understanding of D&I concepts, specifically in AI systems.

**Operationalisation Ability**
3. Participants were asked about their ability to implement D&I principles in their work, with the prompt:
    - "My ability to operationalise D&I in AI principles increased after the workshop." (True/False)
4. **Follow-up**: "How has your ability to implement these principles improved?"
    - The follow-up aimed to uncover how the workshop enhanced participants' practical application of D&I guidelines in product design, governance, or AI model development.

**Team Dynamics**
5. To assess the broader organisational impact, participants were asked about changes in team behaviour:
    - "I have observed changes in the team's approach to considering D&I implications in our use of AI because of this workshop." (True/False)
6. **Follow-up**: "What specific changes have you noticed in your team's approach?"
    - This sought to understand whether the workshop had fostered a shift in collective thinking and collaborative practices regarding D&I.

**Real-World Impact**
7. Finally, participants were asked to evaluate whether the workshop led to tangible improvements in D&I at SEEK:
    - "This workshop contributed to real-world improvements in D&I for SEEK." (True/False)
8. **Follow-up**: "How has the workshop facilitated tangible improvements in D&I at SEEK?"
    - This question aimed to capture specific examples of how the workshop's insights and guidelines translated into measurable changes within SEEK's AI-driven recruitment processes.

### APPENDIX B – EXAMPLES OF PRE-GENERATED USER STORIES FROM D&I IN AI GUIDELINES

These user stories include an identifier and a short phrase describing the D&I in AI guidelines[6]. E.g. P6 refers to the 6th guideline in the "process" pillar, and its short description is: "Create model designs with diversity and inclusion in mind."

**USE CASE 1: Strong Applicant Badge**
**Guideline: [P6: Create model designs with diversity and inclusion in mind]**
- **User story:** As a woman in a male-dominated field, I want the AI matching system to be free from gender biases in its algorithms so that I am judged solely on my qualifications and fit for the role, promoting gender diversity in the industry.
- **User story:** As an international candidate, I want the AI system to recognise and properly evaluate qualifications and experiences from different countries and cultures so that I am not at a disadvantage when applying for jobs that I am qualified for.

**Guideline: [D2: Empower stakeholders and other knowledge holders in data selection, collection, and analysis to ensure demographic representation.]**
- **User story:** As an older jobseeker, I want the AI system to use a data set that includes age representation so that my experience is valued and I'm not excluded due to age-related biases.
- **User story:** As a job placement officer for people with disabilities, I want the AI to be informed by data that includes candidates with disabilities so that the system fairly represents their potential and suggests positions that are accessible and inclusive.
- **User story:** As a female engineer, I want the AI data analysis to be reflective of gender diversity in the tech industry so that the job recommendations I receive are not biased towards traditionally male-dominated roles.

**Guideline: [D8: Enhance feature-based labelling and develop precise user identity notions.]**
- **User story:** As a jobseeker who prioritises gender inclusivity in the workplace, I want the AI's matching algorithm to prioritise companies with a track record of gender diversity so I can work in an environment that values and reflects my identity.

**USE CASE 2: Ad Summarisation**
**Guideline: [P10: Design evaluation tasks that best mirror the real-world setting.]**
- **User story:** As a genderqueer job applicant, I want the AI summarisation process to use gender-neutral language and pronouns in job descriptions so that the language used does not implicitly discourage applicants who do not conform to the gender binary from applying.
- **User story:** As a non-native English speaker, I want the AI to provide job summaries that are clear and free of complex jargon, so that I can understand the role and requirements without being at a disadvantage due to language proficiency.

**Guideline: [H4: Implement inclusive and transparent feedback mechanisms for stakeholders]**
- **User story:** As an ethnic minority professional, I want a straightforward mechanism to report any culturally insensitive language or biases in job summaries so that employers can be more mindful of diversity in their listings and attract a wider range of candidates.
- **User story:** As an AI developer, I want a transparent feedback system for reporting and addressing unintended biases in the AI's language processing so that the job summaries remain fair and inclusive for all user groups.
- **User story:** As a female job applicant in tech, I want a feedback loop in the AI system that allows me to flag when job summaries seem to implicitly skew towards male candidates, ensuring the language used is equally attractive to all genders.
- **User story:** As a professional with a non-traditional career path, I want the AI system to be trained on diverse career trajectories so that the job summaries do not inadvertently favour conventional career progressions over varied experiences.